%% file: video_caption.tex
\documentclass[10pt,twocolumn,letterpaper]{article}

\usepackage{cvpr}
\usepackage{times}
\usepackage{epsfig}
\usepackage{graphicx}
\usepackage{amsmath}
\usepackage{amssymb}

\newcommand {\mymodel}{Memory-Attended Recurrent Network \xspace}

\usepackage{multirow}
\usepackage{booktabs}
\usepackage{tabularx}
\usepackage{arydshln}

\usepackage[pagebackref=true,breaklinks=true,letterpaper=true,colorlinks,bookmarks=false]{hyperref}

 \cvprfinalcopy 

\def\cvprPaperID{3659} 

\ifcvprfinal\fi
\begin{document}

\title{\mymodel for Video Captioning}

\author{Wenjie Pei\textsuperscript{1}, Jiyuan Zhang\textsuperscript{1}, Xiangrong Wang\textsuperscript{2}, Lei Ke\textsuperscript{1}, Xiaoyong Shen\textsuperscript{1} and Yu-Wing Tai\textsuperscript{1}\thanks{Corresponding author is Yu-Wing Tai} \vspace{1mm}\\
\normalsize\textsuperscript{1}Tencent, \quad
	\textsuperscript{2}Southern University of Science and Technology\\
	{\tt \footnotesize wenjiecoder@outlook.com, mikejyzhang@tencent.com, x.wang-2@tudelft.nl\vspace{-1mm}}\\
		{\tt\vspace{-2mm}\footnotesize keleiwhu@gmail.com, goodshenxy@gmail.com, yuwingtai@tencent.com\vspace{-0.05in}}
}


\maketitle


\begin{abstract}
\input{abstract.tex}
\end{abstract}
\vspace{-0.35in}

\section{Introduction}
\input{introduction.tex}

\vspace{-0.1in}
\section{Related Work}
\input{related_work.tex}

\vspace{-0.1in}
\section{\mymodel}
\input{model.tex}

\vspace{-0.1in}
\section{Experiments}
\input{experiments.tex}

\section{Conclusion}
\input{conclusion.tex}

{\small
	\bibliographystyle{ieee}
	\bibliography{egbib}
}

\end{document}

%% file: abstract.tex
Typical techniques for video captioning follow the encoder-decoder framework, which can only focus on one source video being processed. A potential disadvantage of such design is that it cannot capture the multiple visual context information of a word appearing in more than one relevant videos in training data. To tackle this limitation, we propose the Memory-Attended Recurrent Network (MARN) for video captioning, in which a memory structure is designed to explore the full-spectrum correspondence between a word and its various similar visual contexts across videos in training data. Thus, our model is able to achieve a more comprehensive understanding for each word and yield higher captioning quality. Furthermore, the built memory structure enables our method to model the compatibility between adjacent words explicitly instead of asking the model to learn implicitly, as most existing models do. Extensive validation on two real-word datasets demonstrates that our MARN consistently outperforms state-of-the-art methods.

%% file: introduction.tex
\vspace{-0.05in}
Video captioning aims to generate a sequence of words to describe the visual content of a video in a style of natural language. It has extensive applications such as Visual Question Answering (VQA)~\cite{ma2016learning, zeng2017}, video retrieval~\cite{yu2017end} and assisting visually-impaired people~\cite{voykinska2016}. Video captioning is a more challenging problem than its twin `image captioning', which has been widely studied~\cite{anderson2017bottom, rennie2017self, vinyals2015showtell, xu2015showattendtell}. This is not only because video contains substantially more information than still image, but it is also crucial to capture the temporal dynamics to understand the video content as a whole.

\begin{figure}[!tb]
	\begin{center}
		\includegraphics[width=1.0\linewidth]{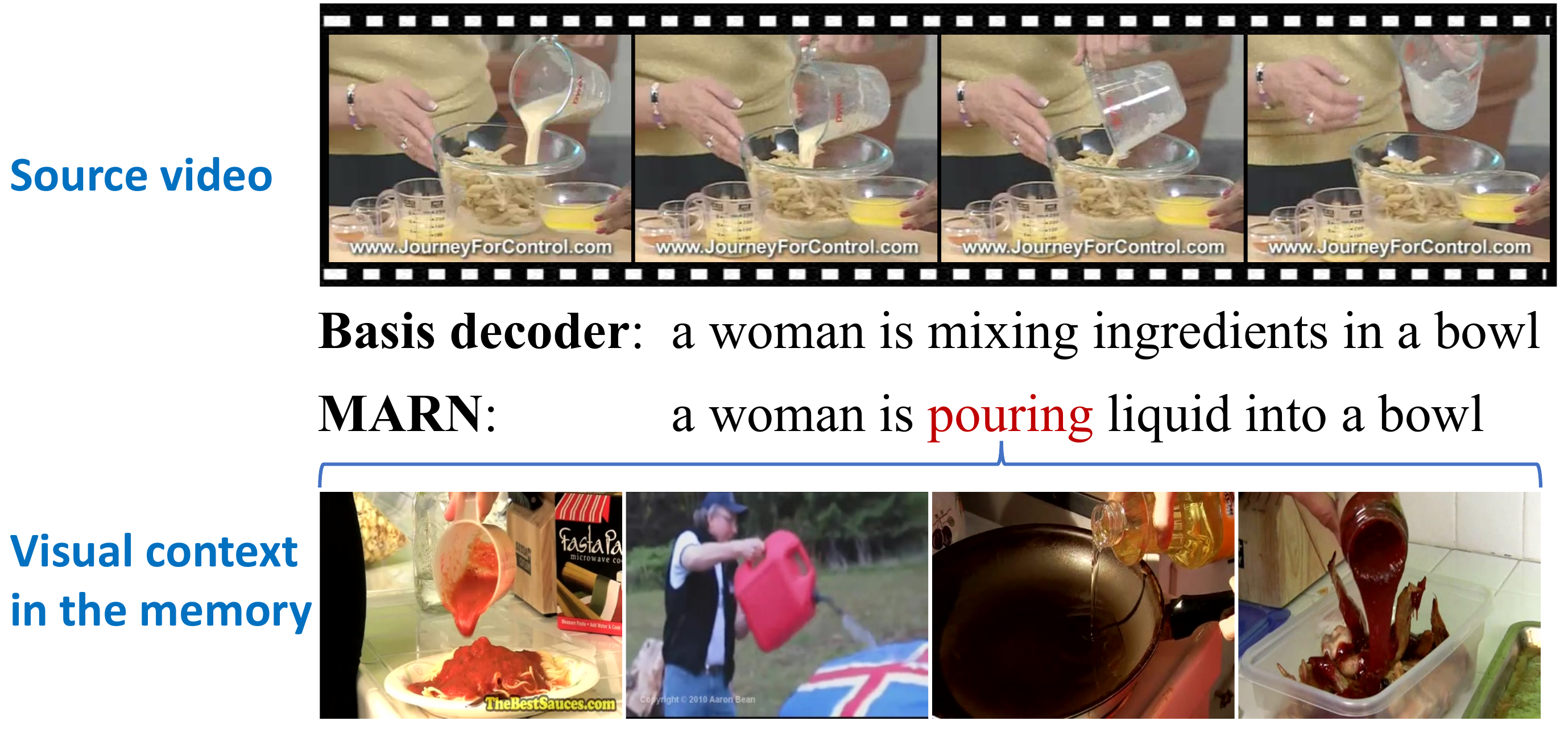}
	\end{center}
	\caption{The typical video captioning models based on the encoder-decoder framework (e.g., the Basis decoder in this figure) can only focus on one source video being processed. Thus, it is hard to explore the comprehensive context information about a candidate word, like `pouring'. In contrast, our proposed MARN is able to capture the full-spectrum correspondence between the candidate word (`pouring' in this example) and its various similar visual contexts (all kinds of pouring actions) across videos in training data 
		, which yields more accurate caption.}
	\label{fig:intro}
\vspace{-0.15in}
\end{figure}

Most existing methods to video captioning follow the  encoder-decoder framework~\cite{gao2017video, jin2016describing, krishna2017dense, liu2017video, pan2016jointly, ramanishka2016multimodal, song2017hierarchical, wang2018_reconstruction, SA-LSTM}, which employs an encoder (typically performed by CNNs or RNNs) to analyze and extract useful visual context features from the source video, and a decoder to generate the caption sequentially. The incorporation of attention mechanism into the decoding process has dramatically improved the performance of video captioning due to its capability of selective focus on the relevant visual content~\cite{gao2017video, krishna2017dense, wang2018bidirectional, SA-LSTM}. One potential limitation of the encoder-decoder framework is that the decoder can only focus on one source video which is currently being processed while decoding. This implies that it can only investigate the correspondence between a word and visual features from a single video input. However, a candidate word in the vocabulary may appear in multiple video scenes that have similar but not identical context information. Consequently existing models cannot effectively explore the full spectrum between the word and its various similar visual contexts across videos in training data. For instance, the basis decoder in Figure~\ref{fig:intro}, which is based on encoder-decoder framework, cannot corresponds the action in the source video to the word `pouring' accurately because of insufficient understanding about the candidate word `pouring'.

Inspired by the memory scheme which is leveraged to incorporate the document context in document-level machine translation~\cite{Gholamreza2018}, in this paper we propose a novel \emph{Memory-Attended Recurrent Network} (MARN) for video captioning which explores the captions of videos with similar visual contents in training data to enhance the quality of generated video caption. Specifically, we first build an attention-based recurrent decoder as the basis decoder,  which follows the encoder-decoder framework. Then we build a memory structure to store the descriptive information for each word in the vocabulary, which is expected to build a full spectrum of correspondence between a word and all of its relevant visual contexts appearing in the training data. Thus, our model is able to obtain a more comprehensive understanding for each word. The constructed memory is further leveraged to perform decoding using an attention mechanism. This memory-based decoder can be considered as an assistant decoder to enhance the captioning quality. 
Figure~\ref{fig:intro} shows that our model can successfully recognize the action `pouring' in the source video because of the full-spectrum contexts (various pouring actions) in the memory.   

Another benefit of MARN is that it can model the compatibility between two adjacent words explicitly. This comes in contrast to the conventional method adopted by most existing models (based on recurrent networks), which  learns the compatibility implicitly by predicting the next word based on the current word and context information.
We evaluate the performance of MARN on two popular datasets (MSR-VTT~\cite{msr-vtt} and MSVD~\cite{msvd}) of video captioning. 
Our model achieves the best results comparing with other state-of-the-art video captioning methods.

%% file: related_work.tex
\vspace{-0.05in}
\paragraph{Video Captioning.}
Traditional video captioning methods are mainly based on template generation which utilizes the word roles (such as subject, verb and object) and language grammar rules to generate video caption. For instance, the Conditional Random Field (CRF) are employed to model different components of a source video~\cite{rohrbach2013translating} and then generate the corresponding caption in a way of machine translation. Also hierarchical structures are utilized to either model the semantic correspondences between concepts of actions and the visual features~\cite{kojima2002natural} or learn the underlying semantic relationships between different sentence components ~\cite{guadarrama2013youtube2text}. Nevertheless, these methods are limited in modeling the language semantics in captions due to the strong dependence on the predefined template. 

As a result of rapid development of deep learning including convolutional networks (CNNs) and recurrent networks (RNNs), the encoder-decoder framework was first introduced by MP-LSTM~\cite{MPlstm}, which employs CNNs as encoder to extract visual features from source videos and then decodes captions by LSTM. Another classical benchmark model based on encoder-decoder framework  is S2VT~\cite{S2VT}, which shares a LSTM in both encoder and decoder. Subsequently, the  attention mechanism gives rise to a significant performance boost to video captioning~\cite{SA-LSTM}. 

Recently, state-of-the-art methods based on encoder-decoder framework seek to make a breakthrough either in the encoding phase~\cite{chen2018less, dong2016early, pan2016hierarchical, ramanishka2016multimodal,  wu2018interpretable} or in the decoding phase~\cite{shetty2016frame, wang2018reconstruction, yu2016video}. Take for examples the cases that focus on the encoding phase,  VideoLAB~\cite{ramanishka2016multimodal} proposes to fuse multiple modalities of source information to improve the captioning performance while PickNet~\cite{chen2018less} aims to pick the informative frames by reinforcement learning. TSA-ED~\cite{wu2018interpretable} proposes to extract the spatial-temporal representation at trajectory level using attention mechanism. In the cases that focus on the decoding phases, RecNet~\cite{wang2018reconstruction} refines the captioning by reconstructing visual features from decoding hidden states and Aalto~\cite{shetty2016frame} designs a evaluator to pick the best caption from multiple candidate captions. 

Most of these methods suffer from a potential drawback that the decoder can only focus on one source video being processed. Hence, they cannot capture the multiple visual context information of a candidate word appearing in rich video context in training data. Our proposed MARN, while following the encoder-decoder framework, is able to mitigate this limitation by incorporating the memory mechanism in the decoding phase to obtain a comprehensive understanding for each candidate word in the vocabulary.


\smallskip\noindent\textbf{Memory-based Models.}
Memory networks were first proposed to rectify the drawback of limited memory of recurrent networks (RNNs)~\cite{NIPS2015_5846, WestonCB14}, which was then extended for various tasks. These memory-based models can be roughly grouped into two categories: 
serves as an assistant module~\cite{D17-1146, Gholamreza2018, Ma2017VisualQA,  xiao2018memory} or dominant module~\cite{P17-2057, jia2018modeling, mohtarami2018automatic, wang2018target}.  In the first category, the memory is leveraged to assist the basis module to enhance the performance of the target task. For instance, two memory components are used to help the basis module, a sentence-based NMT, to capture the document context for document-level machine translation~\cite{Gholamreza2018}. In the second category, the memory serves as a dominant module to perform the target task. An typical example is that memory networks is employed as the backbone for aspect sentiment classification~\cite{wang2018target}. Our proposed MARN falls into the first category since the memory is used as a assistant decoder in our video captioning system. To the best of our knowledge, our MARN is the first to leverage memory network in visual captioning.

%% file: model.tex
\vspace{-0.05in}
\begin{figure*}[htb]
	\begin{center}
		\includegraphics[width=0.9\linewidth]{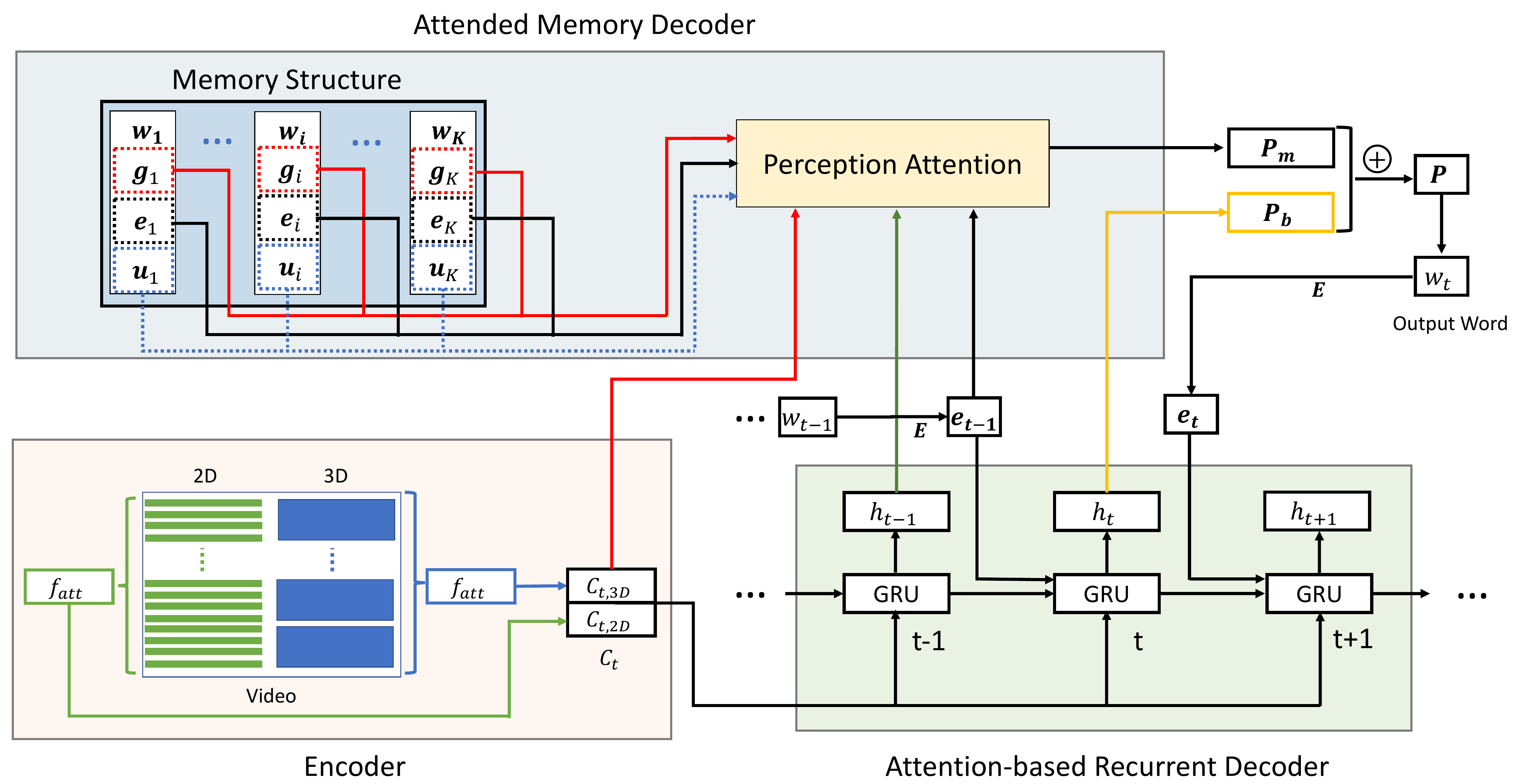}
	\end{center}
	\vspace{-0.1in}
	\caption{The architecture of our \mymodel (MARN). It consists of three components: (1) Encoder for extracting features (both 2D and 3D) from the source video, (2) Attention-based Recurrent Decoder which is used as the basis captioning decoder and (3) Attended Memory Decoder which serves as an assistant decoder to enhance the captioning quality. }
	\label{fig:model}
	\vspace{-0.15in}
\end{figure*}

Our  \mymodel (MARN) consists of three modules: \emph{encoder}, \emph{attention-based recurrent decoder}, and \emph{attended memory decoder}. The overall architecture of MARN is shown in Figure~\ref{fig:model}. After extracting effective features from the source video by the encoder, the attention-based recurrent decoder serves as a basis captioning decoder. Subsequently, the attended memory decoder is designed to enhance the captioning quality as an assistant decoder. We will first introduce the encoder and the attention-based recurrent decoder, then we will elaborate on the proposed attended memory decoder. 

\vspace{-0.05in}
\subsection{Encoder}
\vspace{-0.05in}
\label{sec:encoder}
The role of the encoder is to extract visual features from the input source video, which will be fed to the downstream decoder.  A typical way is to employ pre-trained deep CNNs , such as GoogleNet~\cite{googlenet, SA-LSTM}, VGG~\cite{VGG, msr-vtt} or Inception-V4~\cite{Inception-v4, wang2018reconstruction}, to extract 2D features for each of sampled images from the source video. Similarly, we also rely on the deep CNNs to extract 2D visual features. In our implementation, we opt for the ResNet-101~\cite{resnet} pretrained on imagenet~\cite{imagenet} as the 2D-feature extractor of our encoder due to the its excellent performance and relatively high cost-efficiency. Furthermore, we also extract 3D visual features from the source video to capture the temporal information, which has been shown to be effective in vision tasks involving videos~\cite{krishna2017dense, tran2018closer}. Specifically, we employ the ResNeXt-101~\cite{resnext} with 3D convolutions pretrained on Kinetics dataset~\cite{kinetics} to extract 3D features, which has shown its superior performance on video classification tasks~\cite{hara2018cvpr}. 

Formally, given a sequence of video frames $X = \{x_1, x_2, \dots, x_L\}$ of length $L$, the 2D visual features obtained by pretrained ResNet-101 for each frame are denoted as $F_{2D} = \{ \mathbf{f}_1, \mathbf{f}_2, \dots, \mathbf{f}_L \}$ in which $\mathbf{f}_l \in \mathbb{R}^d$. Besides, the 3D visual features  are extracted by pretrained ResNeXt-101  for every 16 frames, i.e., the temporal resolution for each 3D feature is 16 frames. The resulting 3D features are denoted as $F_{3D} = \{ \mathbf{v}_1, \mathbf{v}_2, \dots,  \mathbf{v}_N \}$, where $N = L / 16$ and $\mathbf{v}_n \in \mathbb{R}^c$. The obtained 2D and 3D visual features are then projected into hidden spaces with the same-dimension $m$ by linear transformations:
\vspace{-0.05in}
\begin{equation}
\mathbf{f'}_l = \mathbf{M}_f \mathbf{f}_l+\mathbf{b}_f, \mathbf{v'}_n = \mathbf{M}_v \mathbf{v}_n+ \mathbf{b}_v.
\label{eqn:encoder1}
\vspace{-0.05in}
\end{equation}
Herein, $\mathbf{M}_f \in \mathbb{R}^{m \times d}$ and $\mathbf{M}_v \in \mathbb{R}^{m \times c}$ are transformation matrices while $\mathbf{b}_f \in \mathbb{R}^m$ and $\mathbf{b}_v \in \mathbb{R}^m$ are bias terms.

\subsection{Attention-based Recurrent Decoder}
The Attention-based Recurrent Decoder is designed as a basis decoder to generate the caption for the source video based on the visual features obtained from the Encoder.  We adopt the similar model structure as Soft-Attention LSTM (SA-LSTM)~\cite{SA-LSTM}. A recurrent neural network is utilized as the backbone of the decoder to generate the caption word by word due to its powerful capability of modeling the temporal information by the recurrent structure. We use GRU~\cite{GRU} in our implementation 
(it is straightforward to replace it with LSTM in our MARN model). 
Meanwhile, the temporal attention mechanism is performed to make the decoder focus on the relevant (salient) visual features when generating each word by automatically learning attention weights for each frame of features.

Concretely, the $t$-th word prediction is performed as a classification task during the decoding process, which calculates the probability of a predicted word $w_k$ among a vocabulary of size $K$ via a softmax function:
\vspace{-0.05in}
\begin{equation}
\vspace{-0.05in}
P_b(w_k) = \frac{\exp\{\mathbf{W}_k \mathbf{h}_t+b_k\}}{\sum_{i=1}^K \exp\{\mathbf{W}_i \mathbf{h}_t+b_i\}},
\label{eqn:softmax1}
\end{equation}
where $\mathbf{W}_i$ and $b_i$ refer to the parameters calculating the linear mapping score for $i$-th word in the vocabulary and $\mathbf{h}_t$ is the learned hidden state of GRU at the $t$-th time. Herein, $\mathbf{h}_t$ is achieved by GRU operations which take into account the hidden state  in the previous step $\mathbf{h}_{t-1}$, the visual context information $\mathbf{c}_t$ and the word embedding of the predicted word in the previous step $\mathbf{e}_{t-1}$:
\vspace{-0.05in}
\begin{equation}
\vspace{-0.05in}
\mathbf{h}_t = \text{GRU}(\mathbf{h}_{t-1}, \mathbf{c}_t, \mathbf{e}_{t-1}),
\end{equation}
where the embedding $\mathbf{e}_{t-1} \in \mathbb{R}^{d'}$ corresponds to the indexed vector in the embedding matrix $\mathbf{E} \in \mathbb{R}^{d' \times K}$.
 The temporal attention mechanism is applied to assign the attention weights for each frame of visual features, including both 2D and 3D features extracted by Encoder. Specifically, the context information of 2D visual features at $t$-th time step is calculated by:
 \vspace{-0.1in}
\begin{equation}
\vspace{-0.1in}
\mathbf{c}_{t, 2D} = \sum_{i=1}^L a_{i,t} \mathbf{f'}_i, \quad a_{i,t} = f_{att} (\mathbf{h}_{t-1}, \mathbf{f'}_i), 
\label{eqn:c2D}
\end{equation}
where $L$ is the length of 2D visual features and $f_{att}$ is the attention function which we adopt the same way as SA-LSTM~\cite{SA-LSTM}: a two-layer perceptron with $\tanh$ activation function in-between. We model the context information of 3D visual features $\mathbf{c}_{t, 3D}$ in the similar way:
\vspace{-0.08in}
\begin{equation}
\vspace{-0.06in}
\mathbf{c}_{t, 3D} = \sum_{i=1}^N a'_{i,t} \mathbf{v'}_i, \quad a'_{i,t} = f_{att} (\mathbf{h}_{t-1}, \mathbf{v'}_i), 
\label{eqn:c3D}
\end{equation}
and we obtain the final context information $\mathbf{c}_t$ by concatenating them together: 
\vspace{-0.08in}
\begin{equation}
\vspace{-0.05in}
\mathbf{c}_t = [\mathbf{c}_{t, 2D}; \mathbf{c}_{t, 3D}].
\end{equation}
We share the attention function $f_{att}$ in both 2D and 3D cases since it is able to guide the optimization of $\mathbf{M}_f$ and $\mathbf{M}_v$ in Equation~\ref{eqn:encoder1} to project both 2D and 3D features into the similar feature space. It can be considered as a regularization to avoid potential overfitting compared to using two independent attention functions. 

It should be noted that our model is different from previous methods utilizing both 2D and 3D visual features in the way how to 
aggregate them. Instead of simply fusing them together by concatenation in the early stage, we treat them separately during encoding and fuse their hidden representations by the attention mechanism in the decoding stage.  A key benefit of this design is that the 2D and 3D features would not be inter-polluted, which is a typical problem as they represent different domains of visual features.
\subsection{Attended Memory Decoder}
We propose the Attended Memory Decoder as an assistant decoder to enhance the quality of the generated caption by the basis decoder (the Attention-based Recurrent Decoder). The rationale behind this design is that a word in the vocabulary may appear in multiple similar video scenes. While the attention-based decoder can only focus on one video scene while decoding, our attended memory decoder is designed to capture the full-spectrum context information from different video scenes where the same candidate word appears and thereby yielding a more comprehensive context for this word. Besides, the conventional attention-based decoder predicts the next word based on current word and context information instead of modeling the compatibility between two adjacent words explicitly. Our Attended Memory Decoder is expected to tackle this issue. 
\vspace{-0.1in}
\subsubsection{Memory Structure}
\vspace{-0.1in}
\label{sec:memory_structure}
The memory is designed to store the descriptive information for each word in the vocabulary. It is constructed as a mapping structure, among which each item is defined as a mapping from a word `$w$' 
to its description `$d$' : $\langle w, d \rangle$.  In particular, the description `$d$' consists of three components: 1) visual context information, 2) word embedding and 3) auxiliary features.

\begin{figure}[!tb]
	\begin{center}
		\includegraphics[width=0.9\linewidth]{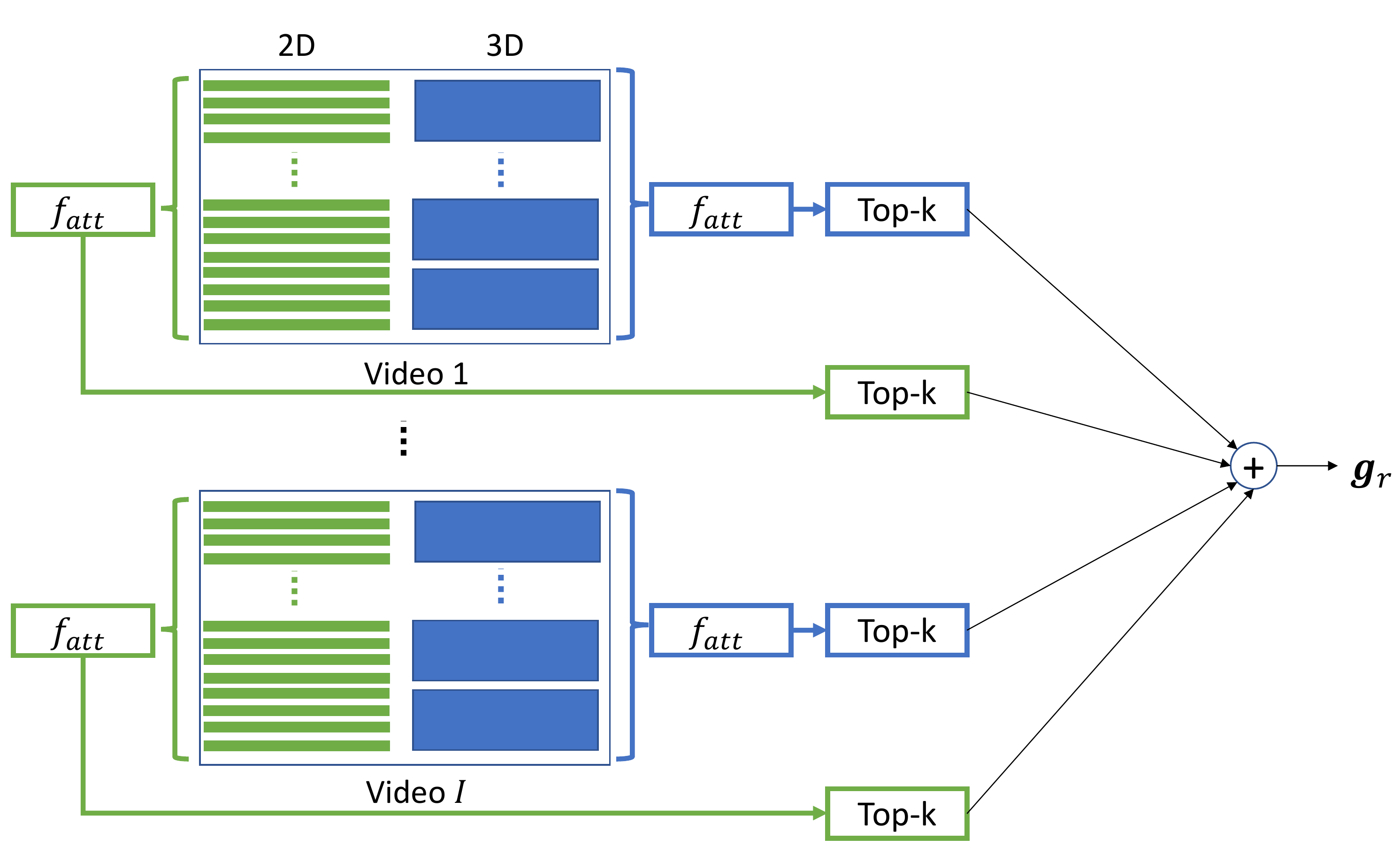}
	\end{center}
	\vspace{-0.1in}
	\caption{The visual context information $\mathbf{g}_r$ for word $w_r$ is constructed by taking into account the top-$k$ relevant frames from each of the related videos. }
	\label{fig:memory}
		\vspace{-0.15in}
\end{figure}

\smallskip\noindent\textbf{Visual context information.}
We extract the visual context information for a given word to describe its corresponding (salient) visual features contained in source videos by attention mechanism similar to Equation~\ref{eqn:c2D} and~\ref{eqn:c3D}. Since a word may appear in multiple video scenes, we extract the salient visual features for each of the videos the word is involved in. To reduce the redundancy of extracted features, we only retain the top-$k$ relevant features for each related video. As shown in Figure~\ref{fig:memory}, the visual context information $\mathbf{g}_r$ for the $r$-th word in the vocabulary is modeled as:
\vspace{-0.08in}
\begin{equation}
\vspace{-0.05in}
\mathbf{g}_r = \frac{\sum_{i=1}^{I} \sum_{j=1}^k (a_{i, j}\mathbf{f'}_{i,j})}{\sum_{i=1}^{I} \sum_{j=1}^k a_{i, j}} + \frac{\sum_{i=1}^{I} \sum_{j=1}^k (a'_{i, j}\mathbf{v'}_{i,j})}{\sum_{i=1}^{I} \sum_{j=1}^k a'_{i, j}},
\end{equation}
where $I$ is the number of related videos to the $r$-th word; $a_{i,j}$ and $a'_{i,j}$ are the $j$-th attention weights among the top $k$ weights for 2D and 3D visual features respectively. Both 2D and 3D context features are normalized to make the magnitude of context features consistent for words with different frequencies. To avoid repetitive modeling, a straightforward way is to train the Attention-based Recurrent Decoder first and then reuse its attention module to extract visual context information for the memory.

\smallskip\noindent\textbf{Word embedding.}
The learned word embedding $\mathbf{e}_r$ of word $w_r$ is also integrated into the memory module to quantitatively describe its properties such as semantics and syntactic features. It is readily achieved once the Attention-based Recurrent Decoder is trained. 

\smallskip\noindent\textbf{Auxiliary features.}
The memory for a word is mainly constructed by the visual context information and word embedding. Additionally, we can also incorporate other potentially-useful auxiliary features, which is denoted as $\mathbf{u}_r$. For instance, we add the category information of videos (when it is available) in the memory, which can help to roughly cluster the video scenes and thereby assisting the decoding process.  

Overall, the memory element corresponding to word $w_r$ is represented as a map structure:
\vspace{-0.08in}
\begin{equation}
\vspace{-0.08in}
\langle w_r, d_r \rangle= \langle w_r, \{ \mathbf{g}_r, \mathbf{e}_r, \mathbf{u}_r \} \rangle.
\end{equation}

\vspace{-0.15in}
\subsubsection{Decoding by Memory}
\vspace{-0.05in}
The constructed memory is leveraged to build a caption decoding system, whose captioning results are further combined with the generated captions by the basis decoder (Attention-based Recurrent Decoder) to improve the captioning quality.

Specifically, we design the memory-attended decoding system as an attention mechanism upon the backbone of attention-based recurrent decoder.  Similar to Equation~\ref{eqn:softmax1}, the probability that word $w_k$ is predicted at the $t$-th time step is modeled via a softmax function:
\vspace{-0.08in}
\begin{equation}
\vspace{-0.08in}
P_m(w_k) = \frac{\exp\{ q_k \}}{\sum_{i=1}^K \exp\{q_i\}},
\label{eqn:softmax2}
\end{equation}
where K is the vocabulary size and $q_i$ is the relevance score for word $w_i$ which is used to measure the qualification of word $w_i$ for the $t$-th time step based on its memory content. There are multiple ways to model the relevance score. We model it  as a simple two-layer perceptron structure: 
\vspace{-0.05in}
\begin{equation}
\vspace{-0.05in}
\begin{aligned}
q_i  = & \ \mathbf{v}^\top \tanh \Big( [\mathbf{W}_c \cdot \mathbf{c}_t + \mathbf{W}_g \cdot \mathbf{g}_i] + [\mathbf{W'}_e \cdot \mathbf{e}_{t-1} + \mathbf{W}_e \cdot \mathbf{e}_i] \\ & + \mathbf{W}_h \cdot \mathbf{h}_{t-1} +  \mathbf{W}_u \cdot \mathbf{u}_i + \mathbf{b} \Big ),
\end{aligned}
\end{equation}
where $\mathbf{c}_t$, $\mathbf{e}_{t-1}$, $\mathbf{h}_{t-1}$ are respectively the context information at time step $t$, predicted word and hidden state at time step $t-1$  from the Attention-based Recurrent Decoder; $\mathbf{W}_c$, $\mathbf{W}_g$, $\mathbf{W'}_e$, $\mathbf{W}_e$, $\mathbf{W}_h$, $\mathbf{W}_u$ are the linear transformation matrices and $\mathbf{b}$ is the bias term.

The physical interpretation behind this modeling is: based on the current situation represented by $\mathbf{h}_{t-1}$, 
the term $[\mathbf{W}_c \cdot \mathbf{c}_t + \mathbf{W}_g \cdot \mathbf{g}_i]$ measures the compatibility between the visual context information of the current source video and the visual context information of the candidate word $w_i$;
$[\mathbf{W'}_e \cdot \mathbf{e}_{t-1} + \mathbf{W}_e \cdot \mathbf{e}_i]$ measures the compatibility between the previously predicted word and the candidate word $w_i$;
the term $ \mathbf{W}_u \cdot \mathbf{u}_r$ corresponds to the auxiliary features.

\smallskip\noindent\textbf{Integrated caption decoding by MARN.}
With Attention-based Recurrent Decoder being the decoding basis and Attended Memory Decoder as the assist, our proposed Memory-Attended Recurrent Network (MARN) models the probability of the word $w_k$ being the next one in the captions as:
\vspace{-0.08in}
\begin{equation}
\vspace{-0.08in}
P(w_k) = (1-\lambda) P_b(w_k) + \lambda P_m(w_k),
\end{equation}
where $\lambda$ is introduced to balance the contribution from two decoders. In practice, the value of $\lambda$ is tuned on a held-out validation set.

\vspace{-0.08in}
\subsection{Parameter Learning}
\vspace{-0.05in}
 Suppose we are given a training set $\mathcal{D} = \{ x^{(n)}_{1, \dots, L^{(n)}}, w^{(n)}_{1, \dots, T^{(n)}} \}_{n=1, \dots, N}$ containing N videos and their associated captions. $L^{(n)}$ and $T^{(n)}$ are respectively the lengths of videos and captions for the $n$-th sample. Since the construction of the Memory relys on the Attention-based Recurrent Decoder, it is trained first and the Attended Memory Decoder is trained subsequently.

 \vspace{-0.12in}
\subsubsection{Attention-based Recurrent Decoder}
\vspace{-0.05in}
Video captioning models are typically optimized by minimizing the negative log likelihood: 
\vspace{-0.07in}
\begin{equation}
\vspace{-0.07in}
L_c = - \sum_{n=1}^N \sum_{t=1}^{T^{(n)}} \log P_b(w^{(n)}_t | x^{(n)}_{1, \dots, L}).
\end{equation}
\vspace{-0.05in}

\smallskip\noindent\textbf{Attention-Coherent Loss (AC Loss)}
\label{sec:acloss}
The visual attention weights learned in Equation~\ref{eqn:c2D}, which is for constructing the context information, always fluctuates significantly even for the adjacent frames since they are learned independently. However, we believe that the attention weights should proceed smoothly. 
Besides, the attention weights, which are assigned to the frames in the time interval corresponding to a event or an action, should be close to each other. It is also consistent with the scheme of human attention. 
To this end,  we propose a so-called Attention-Coherent Loss (AC Loss) to regularize the attention weights in Equation~\ref{eqn:c2D}:
\vspace{-0.05in}
\begin{equation}
\vspace{-0.05in}
L_a = \sum_{n=1}^N \sum_{t=1}^T\sum_{i=2}^L |a^{(n)}_{i,t} - a^{(n)}_{i-1,t}|,
\end{equation}
which minimizes the gap between the attention weights for adjacent frames. Note that the AC Loss is not performed for 3D visual feature because each of the 3D visual features describes the a 3D voxel with a much higher temporal resolution (16 frames in our case) rather than a single frame. Therefore, the smoothness of the attention weights is not required.

Consequently, the Attention-based Recurrent Decoder is trained by minimizing the combined loss:
\vspace{-0.05in}
\begin{equation}
\vspace{-0.05in}
L = L_c + \beta L_a,
\end{equation}
where $\beta$ is a hype-parameter to balance two losses and is tuned on a held-out validation set.
\vspace{-0.15in}
\subsubsection{Attended Memory Decoder}
\vspace{-0.05in}
Similarly, the Attended Memory Decoder is optimized by minimizing the negative log likelihood: 
\vspace{-0.05in}
\begin{equation}
\vspace{-0.1in}
L = - \sum_{n=1}^N \sum_{t=1}^{T^{(n)}} \log P_m(w^{(n)}_t | x^{(n)}_{1, \dots, L}).
\end{equation}

%% file: experiments.tex
\vspace{-0.05in}
We conduct experiments to evaluate the performance of the proposed MARN on two benchmark datasets of video captioning: Microsoft Research-Video to Text (MSR-VTT)~\cite{msr-vtt} and Microsoft Research Video Description Corpus (MSVD)~\cite{msvd}.
We aim to (1) investigate the effect of Attended Memory Decoder on the performance of video captioning and (2) compare our MARN with the state-of-the-art methods for video captioning. 
\vspace{-0.08in}
\subsection{Datasets}
\vspace{-0.05in}
\smallskip\noindent\textbf{MSR-VTT.} MSR-VTT dataset is a widely-used benchmark dataset for video captioning. To have a fair comparison with previous methods, we use the initial version of MSR-VTT, which contains 10,000 video clips from 20 general categories. Each video clip is provided with 20 human-annotated natural sentences (captions) for reference collected by Amazon Mechanical Turk (AMT) workers. We follow the standard data split~\cite{msr-vtt}: 6513 clips for training, 497 clips for test and the left 2990 clips for test.

\smallskip\noindent\textbf{MSVD.} MSVD dataset contains 1970 short video clips collected from YouTube. Each video clip depicts a single activity and is annotated with 40 captions. Following the data split in previous work~\cite{SA-LSTM, Venugopalan_2015_ICCV, wang2018_reconstruction}, 1200 video clips are held out for training, 100 clips for validation and 670 for test.

\vspace{-0.05in}
\subsection{Experimental Setup}
\vspace{-0.05in}
We construct the vocabulary based on the training set by filtering out words occurring fewer than three times, resulting in vocabularies  of around 11K words and 4K words for MSR-VTT and MSVD, respectively. 

The dimension of the word embedding is set to 512. For the GRU in the Attention-based Recurrent Decoder, the number of hidden units is set to 512. For the Encoder, we first extract 2D and 3D features with 2048 dimensions, and then transform them linearly into 512 dimensions as described in Equation~\ref{eqn:encoder1}. The dimensions of the attention modules in the Attention-based Recurrent Decoder and the Attended Memory Decoder are both tuned by selecting the best configuration from the option set $\{256, 384, 512\}$ using a held-out validation set.

We employ Adam~\cite{adam} gradient descent optimization with gradient clipping between -5 and 5~\cite{bengio2013advances}. We perform training for both decoders for 500 epochs with the learning rate decayed by 0.5 every 50 epochs. The final performance is determined by the trained model that performs best on the validation set. To compare our model with state-of-the-art methods, we adopt the standard automatic evaluation metrics, namely CIDEr~\cite{cider}, METEOR~\cite{meteor}, ROUGE-L~\cite{rougel} and BLEU~\cite{bleu}. We use CIDEr, which is especially designed for captioning, as the evaluation metric in our ablation experiments, i.e.,  investigation on the effect of Attended Memory Decoder and Attention-Coherent Loss.

\begin{table}[!thb]
	\centering
	\renewcommand{\arraystretch}{1.3}
	\resizebox{0.8\linewidth}{!}{
		\begin{tabular}[c]{ccc|cc}
			\toprule
			\multicolumn{3}{c|}{Model} & \multicolumn{2}{c}{Dataset}\\
			Basis decoder& $\text{Memory}$ & AC Loss & MSR-VTT & MSVD \\
			\midrule
			$\surd$  &\Large $\times$ \normalsize & \Large $\times$ \normalsize &45.7 & 89.9  \\
			$\surd$ &$\surd$ & \Large $\times$ \normalsize&  46.8 & 91.7  \\
			$\surd$  & $\surd$& $\surd$ & \textbf{47.1} & \textbf{92.2}  \\
			\bottomrule
		\end{tabular}
	}
	\vskip 0.05in
	\caption{Performance measured by CIDEr ($\%$) of our video captioning system equipped with different modules on both MSR-VTT and MSVD datasets (\%) for ablation study. \emph{$\text{Memory}$} refers to the Attended Memory Decoder.}
	\label{table:ablation}
	\vspace{-0.05in}
\end{table}

\begin{table}[!tpb]
	\centering
	\renewcommand{\arraystretch}{1.3}
	\resizebox{1.0\linewidth}{!}{
		\begin{tabular}[c]{ccc|cc}
			\toprule
			\multicolumn{3}{c|}{Memory} & \multicolumn{2}{c}{Dataset}\\
			Word embedding& Visual context & Auxiliary feature & MSR-VTT & MSVD \\
			\midrule
			\Large $\times$ \normalsize & \Large $\times$ \normalsize & \Large $\times$ \normalsize&45.7 & 89.9  \\
			$\surd$  & \Large $\times$ \normalsize& \Large $\times$ \normalsize & 46.1&  90.7\\
			$\surd$ &$\surd$ & \Large $\times$ \normalsize&  46.6 & \textbf{91.7}  \\
			$\surd$  & $\surd$& $\surd$ & \textbf{46.8} & $-$  \\
			\bottomrule
		\end{tabular}
		
		\vspace{-0.15in}
	}
	\vskip 0.05in
	\caption{Performance measured by CIDEr ($\%$) of our video captioning system equipped with different components of the memory on both MSR-VTT and MSVD datasets for ablation study. \emph{$\text{Auxiliary feature}$} refers to the category information in this experiment. Note that AC Loss is not used for all experiments here. The category information is not available for MSVD dataset.}
	\label{table:ablation_memory}
	\vspace{-0.15in}
\end{table}

\vspace{-0.2in}
\subsection{Ablation Study}
\vspace{-0.05in}
We first perform quantitative evaluation to investigate the effect of Attended Memory Decoder and Attention-Coherent Loss respectively. To this end, we conduct ablation experiments which begins with sole basis decoder, namely Attention-based Recurrent Decoder in the captioning system and then incrementally augments the system with Attended Memory Decoder and Attention-Coherent Loss. Table~\ref{table:ablation} presents the experimental results.


\vspace{-0.05in}
\smallskip\noindent\textbf{Effect of Attended Memory Decoder.}
Comparing the performance of sole basis decoder with integrated system of basis decoder and Attended Memory Decoder presented in Table~\ref{table:ablation}, we observe that the Attended Memory Decoder boosts the performance of video captioning by $1.1\%$ and $1.8\%$ (in CIDEr) on MSR-VTT and MSVD, respectively. They are indeed substantial improvements considering the progresses reported in recent years by state-of-the-art methods on video captioning (refer to Table~\ref{table:msrvtt} and~\ref{table:msvd}), which validates the effectiveness of our Attended Memory Decoder.

The memory is composed of three components: visual context, word embedding and auxiliary feature (As explained in Section~\ref{sec:memory_structure}). To further investigate the contribution from each of them to the whole system, we perform ablation study on the memory structure. 
The experimental results presented in Table~\ref{table:ablation_memory} show that the word embedding and the visual context bring about the major performance boost while the auxiliary feature (category information) yields another minor gain on MSR-VTT dataset. 
The word embedding is used for measuring the compatibility between the previously predicted word and current candidate word while the visual context information is responsible for providing a full-spectrum context and measuring how close the candidate word matches the source video. Note that any extra information that is available and potentially helpful for captioning can be readily used as the auxiliary feature.

\smallskip\noindent\textbf{Effect of Attention-Coherent Loss.}
Table~\ref{table:ablation} shows the performance of the system with and without the proposed AC Loss. In particular, the performance is improved by a small margin for both datasets (from $46.8$ to $47.1$ for MSR-VTT and from $91.7$ to $92.2$ for MSVD). 

\vspace{-0.05in}
\subsection{Qualitative Evaluation of Attended Memory Decoder}
\begin{figure*}[!htb]
	\begin{center}
		\begin{tabular}{cccc}
			\includegraphics[width=0.45\linewidth]{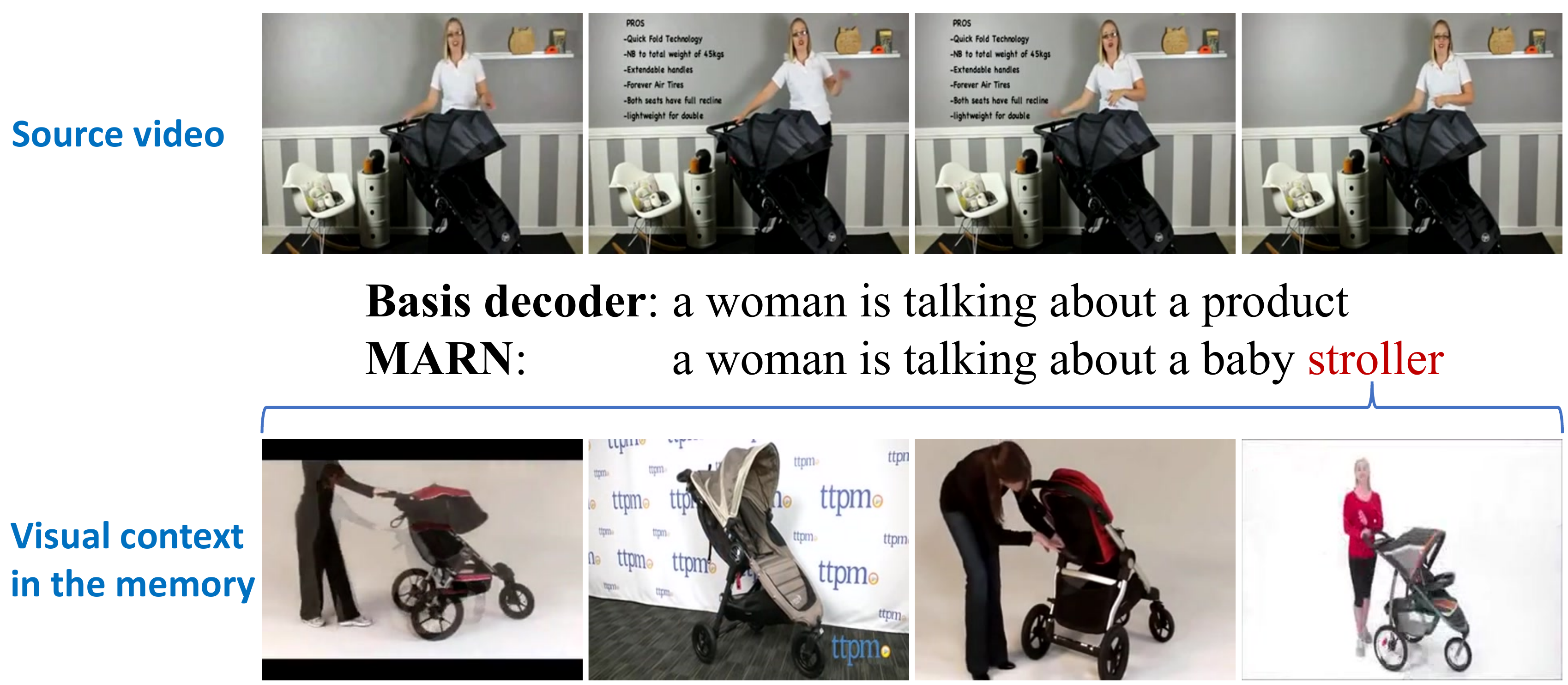} &
			\includegraphics[width=0.45\linewidth]{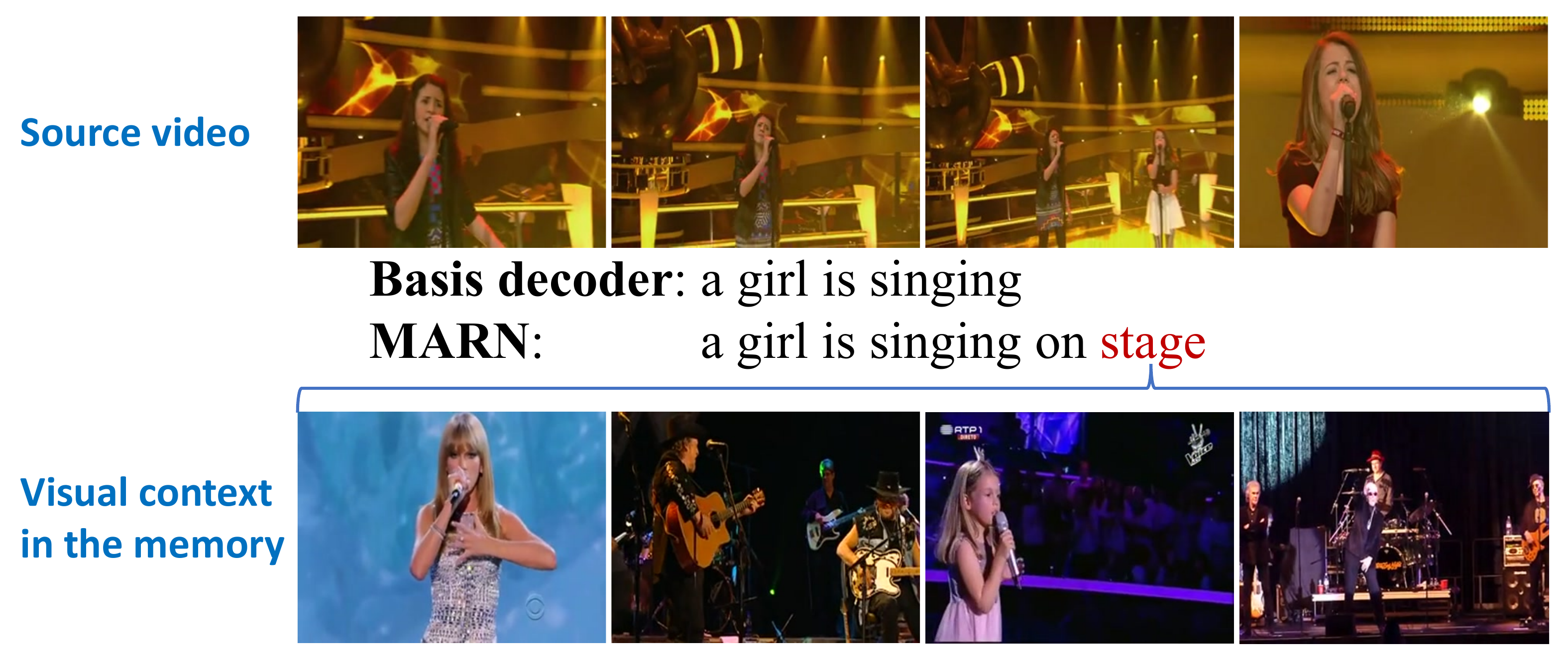}\\
			(a)&(b)\\
			 \includegraphics[width=0.45\linewidth]{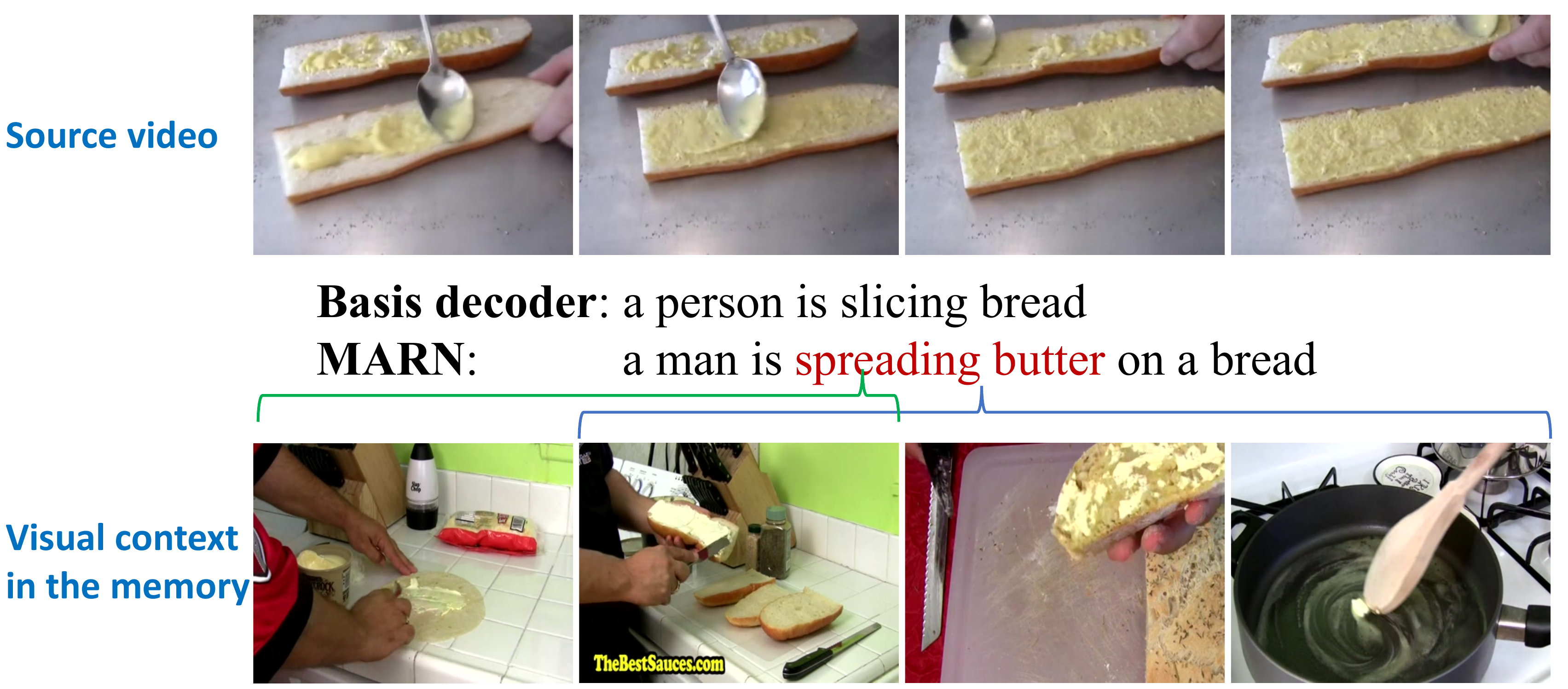} &
			 \includegraphics[width=0.45\linewidth]{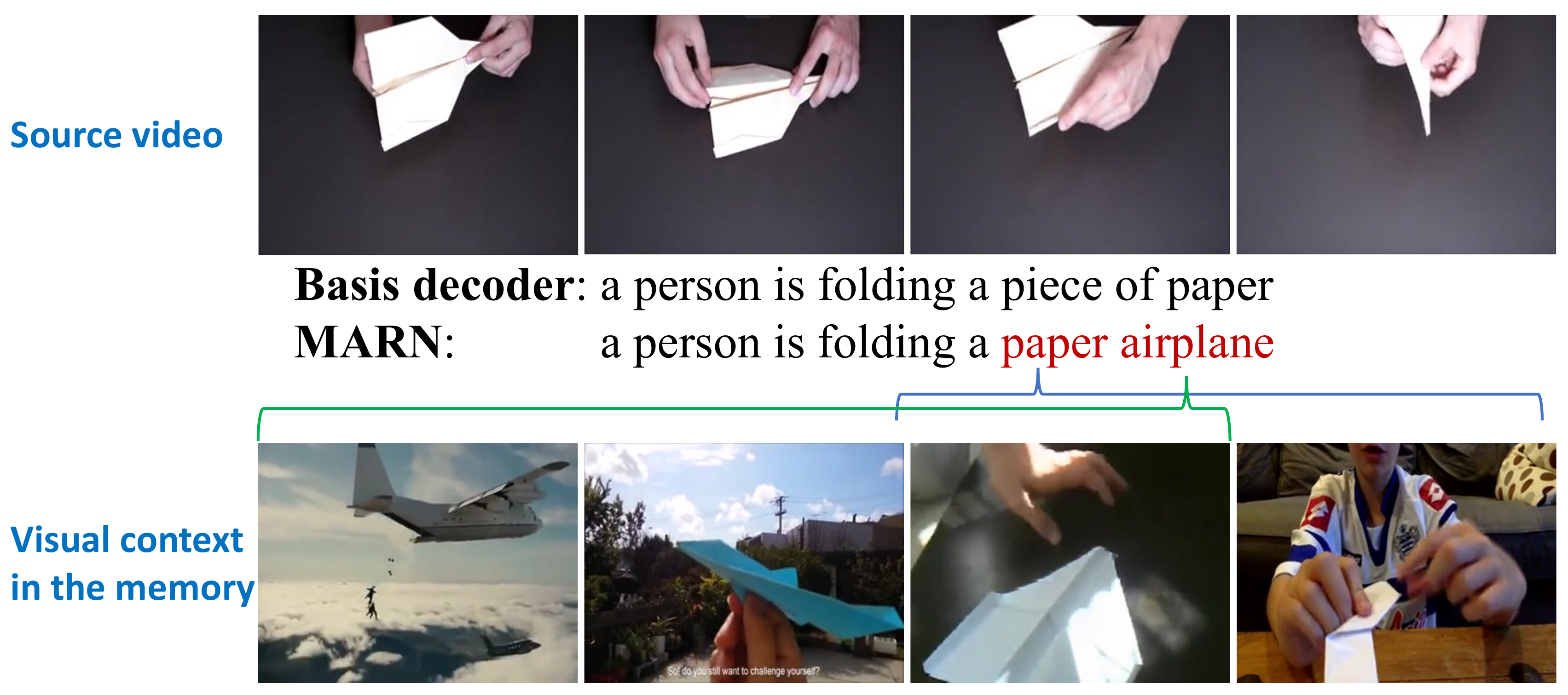}\\
			(c)&(d)\\
		\end{tabular}  
	\end{center}
	\vspace{-0.1in}
	\caption{Qualitative comparison between the basis decoder and our MARN by examples from MSR-VTT and MSVD.  For each example, we first show four representative images of the source video, then we show four context frames (corresponding to the 2D visual contexts in the memory) with high relevance (measured by attention weight) for the key words indicated by red color. MARN is able to correspond the video scene to key words due to the comprehensive understanding of the words by the designed memory scheme. Interestingly, the visual contexts between two adjacent key words in (c) and (d) are overlapped, which may help the model to learn the underlying association. }
	\label{fig:examples}
	\vspace{-0.15in}
\end{figure*}

\vspace{-0.05in}
To gain more insight into what MARN has learned in the memory and the effect of Attended Memory Decoder, we present several examples to qualitatively compare our MARN model with the basis decoder (Attention-based Recurrent Decoder) in Figure~\ref{fig:examples}. Compared to the basis decoder, the MARN is able to decode more precise captions for the given source video, which benefits from the designed Attended Memory Decoder. Take Figure~\ref{fig:examples} (a) as an example, the basis decoder provides a reasonable caption for the video. However, it cannot recognize `baby stroller' while our MARN successfully recognize it due to various `baby stroller' in its memory corresponding to the word `stroller'.



\vspace{-0.05in}
\subsection{Comparison with Other Methods}
\vspace{-0.05in}
Next, we compare our model with existing methods for video captioning on both MSR-VTT and MSVD datasets. All four popular evaluation metrics including CIDEr,  METEOR, ROUGE-L and BLEU are reported.
It should be noted that our model is not compared to the video captioning methods based on reinforcement learning (RL)~\cite{li2019end, wang2018video}, which follows the routine setting in image captioning that the RL-based methods are evaluated separately from other methods (with no RL)~\cite{anderson2017bottom, jiang2018recurrent} for a fair comparison. Nevertheless, it is straightforward to extend our model using RL by applying Self-Critical Sequence Training~\cite{rennie2017self} which is widely adopted in image captioning.
\vspace{-0.12in}
\subsubsection{Comparison on MSR-VTT}
\vspace{-0.08in}
We compare with two groups of baseline methods: 1) fundamental methods including S2VT~\cite{S2VT} which shares a LSTM structure in both encoding and decoding phases, Mean-Pooling LSTM (MP-LSTM)~\cite{MPlstm} which performs a mean-pooling for all sampled visual frames as the input for a LSTM decoder and Soft-Attention LSTM (SA-LSTM)~\cite{SA-LSTM} which employs attention model to summarize visual features for decoding each word; 2) newly published state-of-the-art methods including RecNet~\cite{wang2018reconstruction} which refines the captioning by reconstructing the visual features from decoding hidden states, VideoLAB~\cite{ramanishka2016multimodal} which proposes to fuse source information of multiple modalities to improve the performance, PickNet~\cite{chen2018less} that picks the informative frames based on a reinforcement learning framework, Aalto~\cite{shetty2016frame} that designs a evaluator model to pick the best caption from multiple candidate captions,  and ruc-uva~\cite{dong2016early} which proposes to incorporate tag embeddings in encoding while designing a specific model to re-rank the candidate captions. 

In Table~\ref{table:msrvtt} we show results on MSR-VTT dataset. Our proposed MARN achieves the best performance in terms of METEOR, ROUGE-L and CIDEr while ranking second on BLEU-4. This strongly indicates the superiority of our model. The fact that SA-LSTM outperforms S2VT or MP-LSTM validates the contribution of attention mechanism. Besides, the SA-LSTM equipped with Inception-V4 performs better than its variant with VGG19, which shows the importance of encoding scheme for visual features. The state-of-the-art models typically performs much better than the classical models such as SA-LSTM or MP-LSTM due to all kinds of techniques they proposed. Another interesting observation is that our basis model achieves comparable results with these state-of-the-art models, which somewhat implies the performance ceiling using only encoder-decoder framework and attention mechanism.

\begin{table}[!tpb]
	\centering
	\renewcommand{\arraystretch}{1.3}
	\resizebox{1.0\linewidth}{!}{
		\begin{tabular}[c]{l|cccc}
			\toprule
			Model & BLEU-4 & METEROR & ROUGE-L & CIDEr\\
			\midrule[0.5pt]
			S2VT~\cite{S2VT} & 31.4 & 25.7 & 55.9 & 35.2\\
			MP-LSTM (VGG19)~\cite{wang2018reconstruction} &  34.8 & 24.7 & $-$ & $-$\\
			SA-LSTM (VGG19)~\cite{wang2018reconstruction} & 35.6 & 25.4 & $-$ & $-$\\
			SA-LSTM (Inception-V4)~\cite{wang2018reconstruction} & 36.3 & 25.5 & 58.3 & 39.9\\
			\midrule[0.5pt]
			$\text{RecNet}_{\text{local}}$~\cite{wang2018reconstruction} & 39.1 & 26.6& 59.3 & 42.7\\
			VideoLAB~\cite{ramanishka2016multimodal} & 39.1 & 27.7 & 60.6 & 44.1\\
			PickNet (V+L+C)~\cite{chen2018less} & \textbf{41.3} &27.7 & 59.8 & 44.1\\
			Aalto~\cite{shetty2016frame} & 39.8 & 26.9 & 59.8 & 45.7\\
			ruc-uva~\cite{dong2016early} & 38.7 & 26.9& 58.7 & 45.9\\
			\midrule[0.5pt]
			Basis decoder (ours) & 40.1 & 27.7& 60.4& 45.7  \\
			MARN (ours) & 40.4 & \textbf{28.1} & \textbf{60.7} & \textbf{47.1}\\
			\bottomrule
		\end{tabular}
	}
	\vskip 0.05in
	\caption{Performance of different video captioning models on MSR-VTT dataset in terms of four metrics ($\%$).}
	\label{table:msrvtt}
	\vspace{-0.2in}
\end{table}

\begin{table}[htb]
		\vspace{-0.05in}
	\centering
	\renewcommand{\arraystretch}{1.2}
	\resizebox{0.5\linewidth}{!}{
		\begin{tabular}[c]{p{1cm}<{\centering}|p{1cm}<{\centering}|p{1cm}<{\centering}}
			\toprule
			\multicolumn{3}{c}{MARN vs Basis decoder} \\
			\midrule[0.7pt]
			Win & Tie & Loss \\
			\hline
			43.3$\%$ & 23.3$\%$ & 33.3$\%$\\
			\bottomrule
		\end{tabular}
	}
	\vskip 0.05in
	\caption{Human evaluation for comparing our MARN model with the basis decoder on a subset of MSR-VTT test set. }
	\label{table:human}
	\vspace{-0.15in}
\end{table}

\smallskip\noindent\textbf{Human evaluation.} As a complement to the standard evaluation metrics, we also performs a human evaluation to compare our model with the basis decoder. Specifically, we randomly select a subset from MSR-VTT test sets and ask 30 human subjects to make a comparison between the generated captions by our models and the basis decoder independently. We aggregate evaluation results of all subjects for each sample. Table~\ref{table:human} shows that our model wins among $43.3\%$ test samples and fails on $33.3\%$ samples against the basis decoder, which indicates the advantages of our model.


\vspace{-0.15in}	
\subsubsection{Comparison on MSVD}
	\vspace{-0.1in}
Similar to the experiments on MSR-VTT dataset, two groups of baselines are compared with our model on MSVD dataset: (1) fundamental methods including MP-LSTM with AlexNet as encoding scheme, S2VT and SA-LSTM that both use Inception-V4 for encoding, GRU-RCN~\cite{ballas2015delving} that leverages recurrent convolutional networks to learn video representation, HRNE~\cite{pan2016hierarchical} which proposes a Hierarchical Recurrent Neural Encoder to capture the temporal information of source videos, LSTM-E~\cite{Pan_2016_CVPR} which seeks to explore the decoding with LSTM and visual-semantic embedding simultaneously, 
LSTM-LS~\cite{liu2017video} which aims to model the relationships of different video-sequence pairs, h-RNN~\cite{yu2016video} that employs a paragraph generator to capture the inter-sentence dependency by sentence generators, aLSTMs~\cite{gao2017video} that models both encoder and decoder using LSTM with attention mechanism; (2) three newly published state-of-the-art methods, i.e., PickNet, RecNet and TSA-ED~\cite{wu2018interpretable} which extracts the spatial-temporal representation in the trajectory level by structured attention mechanism.

\begin{table}[thb]
	\centering
	
	\renewcommand{\arraystretch}{1.3}
	\resizebox{1.0\linewidth}{!}{
		\begin{tabular}[c]{l|cccc}
			\toprule
			Model & BLEU-4 & METEROR & ROUGE-L & CIDEr\\
			\midrule[0.5pt]
			MP-LSTM (AlexNet)~\cite{MPlstm} &  33.3 & 29.1 & $-$ & $-$\\
			GRU-RCN~\cite{ballas2015delving} & 43.3 & 31.6 & $-$ & 68.0\\
			HRNE~\cite{pan2016hierarchical} & 43.8 & 33.1 &  $-$ & $-$\\
			LSTM-E~\cite{Pan_2016_CVPR} & 45.3 & 31.0&  $-$ & $-$ \\
			LSTM-LS (VGG19+C3D)~\cite{liu2017video} & 51.1 & 32.6 & $-$ & $-$\\
			h-RNN~\cite{yu2016video}  & 49.9 & 32.6 & $-$ & 65.8\\
			S2VT (Inception-V4)~\cite{wang2018reconstruction} & 39.6 & 31.2 & 67.5 & 66.7 \\
			aLSTMs~\cite{gao2017video} & 50.8 & 33.3 & $-$ & 74.8 \\
			SA-LSTM (Inception-V4)~\cite{wang2018reconstruction} & 45.3 & 31.9 & 64.2 & 76.2 \\
			\midrule[0.5pt]
			TSA-ED~\cite{wu2018interpretable} & 51.7 & 34.0 & $-$ & 74.9\\
			PickNet (V+L)~\cite{chen2018less} & \textbf{52.3} &33.3 & 69.6 & 76.5\\
			$\text{RecNet}_{\text{local}}$(SA-LSTM)~\cite{wang2018reconstruction} & \textbf{52.3} & 34.1& 69.8 & 80.3\\
			\midrule[0.5pt]
			Basis decoder (ours) & 47.5 & 34.4& 71.4& 89.9  \\
			MARN (ours) & 48.6 & \textbf{35.1} & \textbf{71.9} & \textbf{92.2}\\
			\bottomrule
		\end{tabular}
		
	}
	\vskip 0.05in
	\caption{Performance of different video captioning models on MSVD dataset in terms of four metrics ($\%$).}
	\label{table:msvd}
	\vspace{-0.2in}
\end{table}

The experimental results presented in Table~\ref{table:msvd} show that our MARN model performs significantly better than other methods on all metrics except BLEU-4. PickNet and RecNet achieve the best result on BLEU-4. Surprisingly, our basis decoder outperforms other methods substantially, which is mainly beneficial from our encoding scheme, i.e., the combination of 2D and 3D visual features in the specifically-designed way. The performance is further boosted by our Attended Memory Decoder.
\vspace{-0.1in}

%% file: conclusion.tex
\vspace{-0.1in}
In this work, we have presented the Memory-Attended Recurrent Network (MARN) for video captioning. The model employs an attention-based recurrent network as the basis caption decoder and leverages a memory-based decoder to assist the decoding process. The memory is constructed to capture the full-spectrum correspondence between each candidate word and its various visual contexts across videos in training data, which enables the MARN to generate more precise captions for source videos . We show the superior performance of the MARN both quantitatively and qualitatively on two real-world datasets. 